\documentclass{article}
\usepackage{iclr2027_conference,times}

\usepackage{microtype}
\usepackage{booktabs}
\usepackage{graphicx}
\usepackage{amsmath,amssymb}
\usepackage{listings}
\usepackage{xcolor}
\usepackage[hidelinks]{hyperref}
\usepackage{url}
\usepackage{enumitem}
\usepackage[compact]{titlesec}
\titlespacing*{\section}{0pt}{8pt plus 2pt minus 2pt}{4pt plus 1pt minus 1pt}
\titlespacing*{\subsection}{0pt}{7pt plus 2pt minus 2pt}{3pt plus 1pt minus 1pt}
\titlespacing*{\subsubsection}{0pt}{6pt plus 1pt minus 1pt}{3pt plus 1pt minus 1pt}
\titlespacing*{\paragraph}{0pt}{5pt plus 1pt minus 1pt}{6pt}
\setlist[enumerate]{itemsep=1pt,parsep=0pt,topsep=3pt,partopsep=0pt,leftmargin=*}
\setlist[itemize]{itemsep=1pt,parsep=0pt,topsep=3pt,partopsep=0pt,leftmargin=*}
\newcommand{\ci}[2]{{\scriptsize\,$[#1,\,#2]$}}
\newcommand{\code}[1]{\texttt{\small #1}}

\providecommand{\chancanonical}{0.8487}

\providecommand{\chanletters}{0.7971}

\providecommand{\chann}{833}
\providecommand{\channamesonly}{0.8559}

\providecommand{\channeither}{0.5690}

\providecommand{\chanswapnames}{0.1357}

\providecommand{\confaccswap}{0.1357}
\providecommand{\confauccan}{0.8123}
\providecommand{\confaucswap}{0.2433}
\providecommand{\confdropllama}{-0.3235}

\providecommand{\confdropmistral}{-0.3027}

\providecommand{\confdropqwen}{-0.5482}
\providecommand{\confdropqwenabs}{0.548}

\providecommand{\confdroptdabs}{0.713}
\providecommand{\detacc}{0.8921}
\providecommand{\detbase}{0.7420}
\providecommand{\detcov}{0.8100}

\providecommand{\detprec}{0.8977}
\providecommand{\dmislead}{+0.0942}
\providecommand{\dmisleadhi}{+0.1129}
\providecommand{\dmisleadlo}{+0.0754}
\providecommand{\dneutral}{+0.1511}
\providecommand{\dneutralhi}{+0.1646}
\providecommand{\dneutrallo}{+0.1377}
\providecommand{\dneutraln}{3500}
\providecommand{\layaenali}{0.7911}
\providecommand{\layaenmis}{0.5297}
\providecommand{\layaenmislead}{-0.2618}
\providecommand{\layaenmisleadhi}{-0.2386}
\providecommand{\layaenmisleadlo}{-0.2851}
\providecommand{\llamablind}{1.0000}

\providecommand{\lmpolflip}{0.7972}
\providecommand{\mistralneutral}{+0.1531}
\providecommand{\mistralneutralhi}{+0.1724}
\providecommand{\mistralneutrallo}{+0.1343}
\providecommand{\mitqseven}{+0.0748}
\providecommand{\mittdcan}{+0.0391}
\providecommand{\mlchangap}{0.034}
\providecommand{\mthree}{+0.0138}
\providecommand{\mthreehi}{+0.0200}
\providecommand{\mthreelo}{+0.0077}
\providecommand{\nolabcls}{0.8067}
\providecommand{\nolabpol}{0.8914}
\providecommand{\plainneutralseven}{+0.0080}
\providecommand{\plainneutraltd}{+0.1201}
\providecommand{\polcan}{0.6419}
\providecommand{\polcanonical}{0.7420}
\providecommand{\poldigits}{0.8740}
\providecommand{\polflip}{0.9825}
\providecommand{\polfliphi}{0.988}
\providecommand{\polfliplo}{0.977}
\providecommand{\polflipml}{0.9656}
\providecommand{\polletters}{0.8931}

\providecommand{\polneg}{0.3576}

\providecommand{\polunrelated}{0.8537}
\providecommand{\polwarn}{0.3546}

\providecommand{\qsevenperm}{+0.0305}
\providecommand{\qsevenpermhi}{+0.0405}
\providecommand{\qsevenpermlo}{+0.0210}
\providecommand{\qseventbneutral}{+0.1203}
\providecommand{\qseventbneutralhi}{+0.1395}
\providecommand{\qseventbneutrallo}{+0.1019}
\providecommand{\sbaligned}{0.634}
\providecommand{\sbarbitrary}{0.524}

\providecommand{\sevengenMis}{0.8490}

\providecommand{\sevenscoreMis}{0.5010}

\providecommand{\vonchan}{0.8511}
\providecommand{\vonchannodef}{0.5102}
\providecommand{\vonconfauccan}{0.8055}
\providecommand{\vonconfaucswap}{0.8055}

\providecommand{\vonlabclsmis}{0.2281}
\providecommand{\vonlabpold}{+0.0333}
\providecommand{\vonlabpoldhi}{+0.0423}
\providecommand{\vonlabpoldlo}{+0.0246}

\providecommand{\vonpolcan}{0.7382}
\providecommand{\vonpolflip}{0.9711}
\providecommand{\vonpolfliphi}{0.979}
\providecommand{\vonpolfliplo}{0.964}

\providecommand{\xfgenMis}{0.7933}
\providecommand{\xfscoreAli}{0.7886}
\providecommand{\xfscoreMis}{0.8352}

\title{Labels Override Definitions \\ in Jev-Style Typed Decision Models}

\author{Seyedarmin Azizi\thanks{Corresponding author.},\enspace Erfan Baghaei Potraghloo
\& Massoud Pedram \\
University of Southern California (USC), USA \\
\texttt{\{seyedarm, baghaeip, pedram\}@usc.edu}}

\iclrfinalcopy

\begin{document}
\maketitle
\lhead{Preprint}

\begin{abstract}
A typed decision model answers a fixed question about an input by returning a probability for each
of several caller-defined options. Each option carries a short label and a written definition,
which is where a developer states the rule the model should apply. Jev introduced this interface
for routing, moderation and triage, open implementations followed, and the same operation occurs
whenever a language model is used as a classifier by scoring label strings. We study the open
implementations, whose weights we can inspect and patch, and ask whether the probability follows
the definitions or the labels. A preference for the label we call \emph{option-label bias}.
Across four open-weight typed decision models, three ways of reading an answer from a Qwen2.5
backbone, eleven classification tasks and PolicyBench, a synthetic routing suite we introduce in
which the rule appears only in the definitions, the answer is mostly the labels. Deleting every
definition leaves accuracy unchanged (\textsc{laya-td}: \channamesonly{} against
\chancanonical{}), although those definitions support \chanletters{} on their own, and renaming
the options to \code{A} and \code{B} \emph{raises} accuracy by \dneutral{}
\ci{\dneutrallo}{\dneutralhi}. One system, \textsc{von}, is unaffected, and the two code bases
differ in one expression: \textsc{laya} writes each option as
\code{"\{label\}: \{definition\}"}, while \textsc{von} writes only the definition. Changing that
expression in both directions, with no weight changed, makes all three \textsc{laya} checkpoints
exactly invariant ($+0.0000$ $[+0.0000, +0.0000]$) and creates the effect in \textsc{von}, whose
accuracy falls from \vonchan{} to \vonlabclsmis{} when a label contradicts its definition. Earlier
work attributed this failure to the constrained decision head these models use in place of a text
decoder; our results locate it in the prompt rendering. We give a two-call test that tells a
practitioner which case applies to their model, and measure what four mitigations are worth.
\end{abstract}

\begin{figure}[t]\centering
\includegraphics[width=0.92\linewidth]{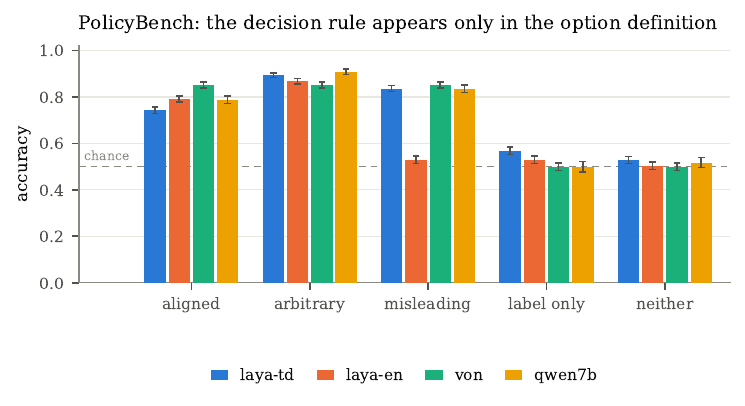}
\caption{PolicyBench, the synthetic routing suite of \S\ref{sec:policybench}, in its controlled
form (\textbf{XF}, in which the two options carry definitions of comparable specificity). The
decision rule is stated only in the option definition, so the label carries no information by
construction and the two right-hand groups are controls. Series are three typed decision models
(\textsc{laya-td}, \textsc{laya-en}, \textsc{von}) and one language-model readout
(Qwen2.5-7B scoring the option strings); \textsc{laya-ml} is near chance on this suite throughout
and is reported in Table~\ref{tab:policy} rather than plotted. Every series except \textsc{von} loses accuracy when the options are given meaningful
names rather than \texttt{A} and \texttt{B}, and
\textsc{laya-en} falls to \layaenmis{} under names that point to the wrong class, close to the
chance level of $0.5$. \textsc{von} is flat across all three label conditions because it never
places the label in the model's input. $n=\dneutraln{}$; error bars are bootstrap 95\% intervals
over items.}
\label{fig:teaser}
\end{figure}

\section{Introduction}
\label{sec:intro}

During 2026 several vendors released small models that do not generate text.
Figure~\ref{fig:teaser} previews the central result. Given an input and a
question, such a model returns a probability for each of a set of options that the caller supplies
with the request. A typical call asks which of three departments should handle a support ticket,
or whether a contract clause contains a particular kind of unfair term. The appeal is cost and
speed: because nothing is generated, one decision costs a single forward pass instead of an
autoregressive decode, and reported costs are one to two orders of magnitude below those of a
frontier language model asked the same question \citep{bench}. These systems are usually called
typed decision models. The first of them, Jev, was released in September 2026 for routing, content
moderation and triage, and open implementations of the same interface followed within days.

Interfaces of this kind ask the caller to supply two things for each option: a short label, such
as \code{billing}, and a written definition, such as ``invoices, payments and refunds''. The
definition is the only place in the request where a developer can state the rule that the model
is supposed to apply. If the rule is ``escalate when the customer is on the Enterprise plan and
asks for a refund'', that sentence goes in the definition field, and the label is a name used to
read the answer back afterwards. The interface therefore implies that the returned probability is
a posterior over the definitions, with the labels serving as identifiers.

Readers who have used a language model as a classifier will recognize the setting. Scoring a fixed
set of label strings under a prompt, and taking the argmax, is the same operation; a typed decision
model is a dedicated model for it, trained to return calibrated probabilities and to do so in one
forward pass. We measure both in this paper, and \S\ref{sec:readouts} states the correspondence
precisely. The effect we report is present in both, so the findings apply to a prompted
classification pipeline as much as to a dedicated model.

This paper tests that implication, and finds that it holds for one of the systems we study and
fails for the others. On three \textsc{laya} checkpoints and on language-model readouts over a
Qwen2.5 backbone, the answer is determined mainly by the labels. Deleting every definition from a
classification request leaves accuracy unchanged, even though the same definitions are sufficient
on their own when the labels are replaced by \code{A} and \code{B}. On a synthetic routing suite
where the rule appears only in the definitions, giving the options meaningful names is worse than
giving them arbitrary ones. A fourth system, \textsc{von}, returns exactly the same probabilities
whatever the labels are.

\citet{sun} reported a related observation on a hosted model, that renaming an option set changes
its answers, and attributed the effect to the constrained decision head that these models use in
place of a decoder. Our measurements do not support that attribution. An ordinary language model
scoring the same option strings shows the effect at least as strongly, and the effect can be
removed from a typed decision head, or added to one that does not have it, by changing how the
options are written into the input and nothing else.
What separates the systems that are affected from the one that is not is visible in the released
code: whether the option label is written into the model input alongside its definition. We
establish this by changing that one expression in both directions rather than by comparing the two
systems, so the result does not depend on their other differences.

An evaluation audit of the first twenty-eight papers on these models counted how many applied two
basic checks. Of those for which each check was applicable, only 5 of 26 included a
label-probability baseline from an ordinary language model, and only 1 of 27 tested whether a
decision is invariant to the names given to the options
\citep{audit}. The two
omissions are related, because without the invariance test the effect is not visible, and without the
baseline it cannot be attributed to anything in particular.

\paragraph{Contributions.}
\begin{enumerate}
\item An ablation that separates the two channels a request carries, the option label and the
      option definition, and measures what each contributes (\S\ref{sec:channels}).
\item \textbf{PolicyBench}, a synthetic routing suite in which the decision rule appears only in
      the option definitions, the correct answer is computed from the generating attributes, and
      the labels are varied from aligned through arbitrary to misleading
      (\S\ref{sec:policybench}, \S\ref{sec:policyresults}).
\item An intervention in both directions on two independently trained systems: removing the
      option label from the input of three \textsc{laya} checkpoints eliminates the effect, and
      adding it to \textsc{von} creates it, with no change to weights, decision head or
      calibration (\S\ref{sec:intervention}).
\item Evidence against the published attribution of this failure to the constrained decision
      head, from a language-model readout over the same option strings and from three readouts
      over one fixed backbone (\S\ref{sec:lm}).
\item A two-call test that tells a caller which regime a given model is in, requiring no labels,
      no held-out data and no access to weights (\S\ref{sec:diagnostic}).
\item Measurements of a second and independent failure, in which a negated yes/no question is not
      read at all (\S\ref{sec:polarity}), and of the behavior of the models' own confidence
      scores when a decision has been corrupted (\S\ref{sec:confidence}).
\item Four mitigations, each measured rather than recommended on principle
      (\S\ref{sec:mitigations}).
\end{enumerate}

\paragraph{What to do about it.}
A reader who uses one of these interfaces can act on three findings without reading the rest of
the paper. First, run the two-call test of \S\ref{sec:diagnostic}: ask one question twice,
changing only the option labels and leaving the definitions alone, and compare the returned
distributions. If they differ, the labels are affecting the decision. Second, if they differ and
the rule you care about lives in the definitions rather than in the class names, rename the
options to \code{A}, \code{B} and so on; on our routing suite this is worth \dneutral{} accuracy
\ci{\dneutrallo}{\dneutralhi}. Third, do not rely on a confidence threshold to catch this: when a
label and its definition disagree, the confidence score of the models we tested ranks their errors
in the wrong direction (\S\ref{sec:confidence}). Section~\ref{sec:mitigations} gives the measured
value of four mitigations, including what to do when the labels have to keep their meaning.

\paragraph{Scope.}
Every model we study has open weights, and every number in this paper can be reproduced without a
vendor API key. We do not measure Jev itself. The audit cited above notes that
its ``internals remain closed, so claims about mechanism apply only to open systems''
\citep{audit}, and the claims we make are about mechanism. The open implementations are also what
is deployed locally by a substantial part of this ecosystem \citep{wild}.

\section{Background}
\label{sec:background}

\subsection{What a typed decision model does}
A request consists of a \emph{state}, which is the input to be judged, and one or more
\emph{questions}. Each question has a type, an instruction, and, for most types, a set of options
with definitions. The model returns one answer per question. No text is produced, and the set of
possible answers is fixed by the request, so the output can be parsed without error handling.
Three question types are common to the interfaces we study:

\textbf{choice} selects one of a set of named options, and the reply contains the selected option,
a probability for each option, and a confidence score. \textbf{score} places the state on an
ordered scale of named levels, and the reply contains a numeric position and a distribution over
levels. \textbf{boolean} judges whether a statement about the state is true, and the reply is a
single probability.

Figure~\ref{fig:example} is a complete \code{choice} call to \code{laya-typed-decisions} with the
reply it produces. The \code{criteria} field carries the definitions.

\begin{figure}[ht]
\begin{lstlisting}[language=Python]
state = ("Hi, we were billed twice for March. Please refund the duplicate "
         "today or we will cancel our plan.")
question = {"type": "choice",
            "instructions": "Which department should handle this ticket?",
            "criteria": {"billing":   "invoices, payments and refunds",
                         "technical": "bugs, outages and system errors",
                         "other":     "anything else"}}

# reply
{"choice": "billing",
 "probabilities": {"billing": 0.8436, "technical": 0.0803, "other": 0.0761},
 "confidence": 0.5067}
\end{lstlisting}
\caption{A complete request and reply. The caller supplies a state and one typed question; the
model returns a selected option, a probability for each option, and a confidence score. The
\code{criteria} field is where the meaning of each option is given.}
\label{fig:example}
\end{figure}

In this example the label \code{billing} and the definition bound to it both point to the same
answer, which is the usual situation and the reason the distinction is easy to overlook. The
experiments in this paper separate the two.

\subsection{The two channels of a request}
\label{sec:channels_def}
Each option in a \code{choice} request reaches the model through two pieces of text: its
\emph{label}, such as \code{billing}, and its \emph{definition}, such as ``invoices, payments and
refunds''. We call these the two \textbf{channels} of the request. Either one on its own can
indicate which option is correct. The label does so when it is an ordinary name for the class, as
\code{billing} is for a billing ticket. The definition does so when it states the rule, which is
the case the interface is designed for and the only case available when the rule is something a
pretrained model could not already know. Throughout the paper we vary the two channels
independently and report what each contributes.

\subsection{How an answer is obtained without generating text}
\label{sec:readouts}
We use \emph{readout} for the procedure that turns one forward pass into a probability over the
caller's options. Write $s$ for the state, $q$ for the question, and $o_1,\dots,o_k$ for the
options. Three readouts appear in this paper, and the difference between them is what
\S\ref{sec:lm} rests on.

A \textbf{typed decision head} inserts a marker token into the input for each option, reads a
scalar $z_i$ from the encoder at the $i$-th marker position, and normalizes across markers:
\begin{equation}
p(o_i \mid s, q) \;=\; \frac{\exp(z_i / \tau)}{\sum_{j=1}^{k}\exp(z_j / \tau)} ,
\label{eq:head}
\end{equation}
with $\tau$ a calibration temperature. All $k$ scores come from one pass, which is what makes a
decision cost a single forward pass. This is the readout used by \textsc{laya} and \textsc{von}.

A \textbf{label log-probability readout} uses an ordinary language model. It renders $s$ and $q$
into a prompt $\pi(s,q)$ and scores each option string as a continuation, normalizing by the
number of tokens $|o_i|$ so that longer option names are not penalized:
\begin{equation}
\ell_i \;=\; \frac{1}{|o_i|}\log P_{\mathrm{LM}}\!\left(o_i \mid \pi(s,q)\right),
\qquad
p(o_i \mid s, q) \;=\; \frac{\exp \ell_i}{\sum_{j=1}^{k} \exp \ell_j}.
\label{eq:logprob}
\end{equation}
A \textbf{first-token readout} is the same construction restricted to the first token at which the
options differ, so that $\ell_i = \log P_{\mathrm{LM}}(o_i^{(1)} \mid \pi(s,q))$. Both are standard
ways of using a language model as a classifier.

Equations~\eqref{eq:head} and~\eqref{eq:logprob} share a property that matters here: the answer is
obtained by comparing the option strings with one another. We call these \textbf{scoring readouts}.

A \textbf{generative readout} instead samples text from the model and matches the reply back to an
option:
\begin{equation}
\hat{o} \;=\; \operatorname{match}\!\big(\operatorname{decode}(\pi(s,q)),\; \{o_1,\dots,o_k\}\big).
\label{eq:gen}
\end{equation}
It returns a decision rather than a distribution, and it does not compare option strings against
one another. Holding the model fixed and switching between \eqref{eq:logprob} and \eqref{eq:gen}
is what isolates the second half of our result in \S\ref{sec:lm}.

\subsection{Systems studied}
\textsc{laya} and \textsc{von} are open-weight implementations of the Jev interface: they accept
the same request format, expose the same \code{criteria} field, and return the same three answer
types. We study them rather than Jev itself because the mechanism we identify is a property of how
a request is rendered into the model's input, which can only be read and patched in a system whose
code and weights are available. \textsc{laya} is a family built on ModernBERT
\citep{modernbert} and released under Apache-2.0. We use three checkpoints, and refer to each by
the short name used in every table and figure: \code{laya-typed-decisions} (421M parameters,
written \textsc{laya-td}), \code{laya} (421M, written \textsc{laya-en}) and
\code{laya-multilingual} (322M, written \textsc{laya-ml}). \textsc{von} is a separate open implementation, also built on ModernBERT, with 395M
parameters. Where a
statement applies to all three \textsc{laya} checkpoints we say so explicitly; an unqualified
short name always means that one checkpoint. For the language-model readouts we use Qwen2.5 at 0.5B, 1.5B and 7B parameters, and
Llama-3.1-8B-Instruct and Mistral-7B-Instruct-v0.3 to check that the result is not specific to one
pretraining run.

One architectural detail is relevant later, which is that \textsc{laya} constructs one input sequence per
question, so a request carrying several questions is a batch of sequences rather than a single
shared context. Questions in one request therefore cannot influence one another, and none of the
effects we report can be explained that way.

\section{Related work}
\label{sec:related}

\paragraph{Evaluations of typed decision models.}
A body of evaluation work appeared within weeks of Jev's release, most of it evaluating the
hosted model. \citet{bench} report
accuracy across a wide set of classification and reading-comprehension tasks. \citet{axioms} show
that the returned probabilities violate the probability axioms when logically linked questions
are asked. \citet{jevout} change decisions by adding naturally phrased context, \citet{jevadv}
by appending an unverified opinion, and \citet{hijack} by prompt injection. \citet{wild} survey
public projects built on these models. \citet{audit} review this literature and report the two
evaluation gaps quoted in \S\ref{sec:intro}.

\paragraph{The effect we study.}
\citet{sun} renamed option sets and reported large changes in the answers of Jev, the hosted
model, which they attributed to the constrained decision head. We reproduce the effect on open weights, measure it
against a language-model baseline that the earlier work did not include, and locate it in the
prompt rendering rather than in the decision head, by changing that rendering in both directions. We also separate it from a second failure,
insensitivity to question polarity, which is present in every typed model we test including the one
that is invariant to labels.

\paragraph{Input-independent label priors.}
That a language model's label probabilities carry a large input-independent prior is established,
and contextual calibration removes it by dividing out the prediction on a content-free input
\citep{zhao2021}. We evaluate that correction here and report the conditions under which it helps
and the conditions under which it does not (\S\ref{sec:mitigations}). That prior and the effect we
study are distinct. An input-independent prior is a tendency to return one option whatever the
state is, and it is present even when the options carry no definitions. Option-label bias is a
tendency to answer from an option's name in preference to the definition supplied for it, and it
is only defined when the two channels of \S\ref{sec:channels_def} can disagree. Equation
\eqref{eq:blind} measures the first; the contrast between aligned and arbitrary labels measures
the second.

\section{Method}
\label{sec:method}

\subsection{Varying the two channels}
\label{sec:variants}
A \emph{variant} is a rewriting of the same request that changes how the options are presented
while leaving what they mean unchanged. In every variant, the definition attached to a presented
label still correctly describes the class that label stands for, so a model that answers from the
definitions is unaffected by any of them. We record, for each variant, which presented label
carries the definition of each true class, so a class probability is always read back by the
right name.

\begin{table}[ht]\centering\small
\noindent\resizebox{\linewidth}{!}{\begin{tabular}{llll}
\toprule
variant & labels & definitions & what it isolates \\
\midrule
\code{canonical} & the task's own class names & correct & the deployed configuration \\
\code{letters} & \code{A}, \code{B}, \dots & correct & the definition channel alone \\
\code{letters\_perm} & \code{B}, \code{A}, \dots & correct & definition following, no contradiction \\
\code{unrelated} & \code{alpha}, \code{beta}, \dots & correct & whether neutrality needs a letter \\
\code{digits} & \code{0}, \code{1}, \dots & correct & as above \\
\code{synonym} & close synonyms & correct & a low dose of mismatch \\
\code{swap\_names} & each class under another's name & correct & label/definition conflict \\
\code{order} & canonical & correct, shuffled order & presentation position \\
\code{names\_only} & canonical & stripped to a placeholder & the label channel alone \\
\code{neither} & \code{A}, \code{B}, \dots & stripped & control: no usable channel \\
\code{warned} & canonical, plus a warning & correct & is the failure instruction-addressable \\
\code{swap\_warned} & swapped, plus a warning & correct & as above, under conflict \\
\code{negated} & yes/no only, polarity inverted & -- & whether polarity is read \\
\bottomrule
\end{tabular}
}
\caption{The variants used throughout. The first column gives the identifier used in the released
code and in the raw result files; the plain names introduced below are used in all prose, tables
and figures.}
\label{tab:variants}
\end{table}

Table~\ref{tab:variants} lists every variant with the identifier used in the released code. We
use one name for each throughout the paper, and the tables use the same names:

In \textbf{aligned} the options keep the task's own class names, which is the ordinary way a
request is written. In \textbf{arbitrary} the names are replaced by \code{A}, \code{B} and so on,
which carry no meaning; only the definitions are informative, so this condition also measures the
definition channel on its own. \textbf{Arbitrary, exchanged} uses the same arbitrary names with
the definitions attached the other way round. In \textbf{misleading} each class is presented under
a different class's name, while its definition still describes it correctly. In \textbf{label
only} the definitions are replaced by a placeholder, leaving the labels as the sole informative
channel. In \textbf{neither} both channels are stripped; this is the no-information control, and a
model should be at chance here.

The comparison between \textbf{arbitrary} and \textbf{arbitrary, exchanged} carries most of the
argument, because neither request contains a contradiction. Both present labels with no meaning of
their own alongside correct definitions, so the correct class is unambiguous in each, and a model
that answers from the definitions scores the same under both.

We vary the state as well, so that in the \textbf{real} condition the model receives the actual input; in
\textbf{null} it receives an empty state, which also supplies the prior used in
\S\ref{sec:mitigations}; in \textbf{mismatch} it receives another item's state, which checks
whether the model is using \emph{this} input rather than responding to the question alone.

\subsection{Classification tasks}
\label{sec:tasks}
Eleven public tasks supply the ordinary classification setting. Three of them are
\emph{fixed-label choice} tasks, meaning every item is a \code{choice} question over the same
option set, so the labels and the definitions can be varied independently across the whole task:
SST-2 \citep{sst}, RTE and CB \citep{superglue}. Seven are yes/no tasks used for the polarity
measurement of \S\ref{sec:polarity}: BoolQ \citep{boolq}, WiC \citep{superglue}, yes/no forms of
SST-2 and RTE, and three clause types from UNFAIR-ToS \citep{lexglue}. One is ordinal, SST-5
\citep{sst}, and is used only where an ordered scale is relevant. Per-task sizes and class
prevalence are given in Appendix~\ref{app:tasks}.

\subsection{PolicyBench}
\label{sec:policybench}
On an ordinary classification task the label already names the answer, so a model can score well
without reading the definition at all. PolicyBench is constructed so that this is not possible.
Support tickets are generated from templates
over six controlled attributes. Each policy is a boolean function of those attributes, stated in
natural language \emph{only} in the definition bound to the true branch. Ground truth is computed
from the generating attributes, so there is no label noise, no annotator disagreement and no
possibility of contamination, and classes are exactly balanced by rejection sampling. Seven
policies span single-term, conjunctive, disjunctive and negated rules.

The label conditions of \S\ref{sec:variants} apply directly. In the aligned condition the
options are named \code{escalate} and \code{hold}; in the arbitrary condition they are named
\code{A} and \code{B}; in the misleading condition the option whose definition is the policy is
named \code{hold} and the other is named \code{escalate}.

A control suite, \textbf{PolicyBench-XF} (for \emph{explicit false branch}), replaces the false
branch's catch-all definition (``none of the above applies'') with an explicit negation of the
policy, so that the two options carry
definitions of comparable substance. \S\ref{sec:confound} shows why this control is necessary.
We report XF as the primary suite throughout.

\subsection{Models}
\label{sec:models}
Typed decision heads: \code{laya-typed-decisions} (421M), \code{laya} (421M),
\code{laya-multilingual} (322M), and \code{von} (395M). Readouts over one Qwen2.5 backbone,
which separates the typed head from the task format: length-normalized label log-probability,
constrained first-token, and generate-then-parse, at 0.5B, 1.5B and 7B.

\subsection{Metrics}
\label{sec:metrics}
We report accuracy throughout, and AUROC where a confidence score is being evaluated as a
ranking of the model's own correctness (\S\ref{sec:confidence}). Two further statistics are
specific to this paper.

The \textbf{flip rate} between two variants $u$ and $v$ is the fraction of items on which the
decision changes, where the decision is the highest-probability class:
\begin{equation}
\mathrm{flip}(u,v) \;=\; \frac{1}{n}\sum_{m=1}^{n}
\mathbf{1}\!\left[\arg\max_i p_u(o_i \mid s_m) \neq \arg\max_i p_v(o_i \mid s_m)\right].
\label{eq:flip}
\end{equation}
It measures whether a rewriting changes what the model decides, independently of whether the
decision was correct.

The \textbf{state-blindness} of a variant is the fraction of items on which the model's decision
equals the decision it makes for the same request with an empty state:
\begin{equation}
\mathrm{blind}(u) \;=\; \frac{1}{n}\sum_{m=1}^{n}
\mathbf{1}\!\left[\arg\max_i p_u(o_i \mid s_m) = \arg\max_i p_u(o_i \mid \varnothing)\right].
\label{eq:blind}
\end{equation}
A value near $1$ means the model is returning nearly the same answer whatever the input is, so a
high accuracy in that condition would have to come from the class balance rather than from
reading the state.

All intervals are percentile bootstrap over items with 4000 resamples, paired when two conditions
are compared on the same items. Per-task sizes and class prevalence are in
Appendix~\ref{app:tasks}.

\section{Results}
\label{sec:results}

\subsection{What each channel contributes}
\label{sec:channels}
\begin{figure}[t]\centering
\includegraphics[width=\linewidth]{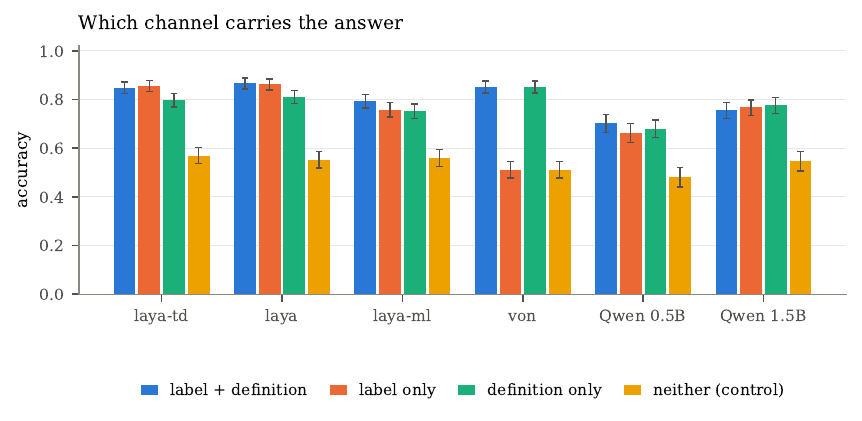}
\caption{What each channel contributes, on the fixed-label choice tasks (SST-2, RTE, CB;
$n=\chann{}$). For every model except \textsc{von}, supplying only the labels (second bar) is as
accurate as supplying both channels (first bar), so the definitions add nothing when a meaningful
label is present. \textsc{von} shows the opposite pattern: it matches its full accuracy from the
definitions alone and falls to chance when they are removed. Exact values with intervals are in
Table~\ref{tab:channels} of Appendix~\ref{app:tables}. Error bars are bootstrap 95\% intervals.}
\label{fig:channels}
\end{figure}

Figure~\ref{fig:channels} reports the ablation. For \code{laya-typed-decisions}, deleting every
definition does not reduce accuracy (\channamesonly{} against \chancanonical{}, intervals overlapping), and the same holds
for \textsc{laya-en}. For \textsc{laya-ml} the two conditions differ by \mlchangap{} with intervals that
overlap. The definitions are not unreadable to the model, because when they are supplied as the
only informative channel, with the labels replaced by letters, they support \chanletters{}, well above
the \channeither{} control in which neither channel carries information. The definitions are
therefore legible to the model and are not being used when a label is available. When a label and
its definition point to different classes, accuracy falls to \chanswapnames{}, below the
no-information control, which means the decision is following the label.

The \textsc{von} row behaves differently, in that its accuracy is \vonchan{} whether the labels are
the class names, arbitrary letters, or the names of the wrong classes, and \vonchannodef{} in both
conditions where the definitions have been removed. Because \textsc{von} does not use the labels,
these three requests are the same input to it. When the definitions are removed it has no
information left and performs at chance, whereas \textsc{laya} still reaches \channamesonly{} from
the labels alone. Together these two observations describe a trade: reading only the definitions
costs nothing in accuracy on this suite (\vonchan{} against \chancanonical{}), but it requires the
caller to have written definitions that are sufficient on their own.

\paragraph{The \code{score} primitive admits no such ablation.}
A \code{choice} question carries labels and definitions in separate fields, which is what makes
this ablation possible. A \code{score} question does not: its \code{criteria} is the ordered list
of level labels, so the label and the definition are the same string and there is no second channel to remove.
We therefore exclude the ordinal task from this ablation. This is a property of the interface rather than of any model. It has a consequence for callers:
a rubric level cannot be given a definition that its label does not already carry, so the guidance
of \S\ref{sec:mitigations} for choosing neutral labels cannot be applied to score questions at
all.

\subsection{Results on PolicyBench}
\label{sec:policyresults}
\begin{table}[t]\centering\small
\noindent\resizebox{\linewidth}{!}{\begin{tabular}{lrrrrrrr}
\toprule
model & aligned & arbitrary \texttt{alpha}/\texttt{beta} & arbitrary \texttt{0}/\texttt{1} & arbitrary \texttt{A}/\texttt{B} & misleading & label only & neither \\
\midrule
\textsc{laya-td} & 0.742 \ci{0.728}{0.756} & 0.854 \ci{0.842}{0.865} & 0.874 \ci{0.863}{0.885} & 0.893 \ci{0.883}{0.903} & 0.836 \ci{0.824}{0.849} & 0.569 \ci{0.553}{0.585} & 0.528 \ci{0.512}{0.545} \\
\textsc{laya-en} & 0.791 \ci{0.778}{0.804} & 0.836 \ci{0.823}{0.848} & 0.846 \ci{0.834}{0.858} & 0.869 \ci{0.857}{0.880} & 0.530 \ci{0.513}{0.546} & 0.529 \ci{0.513}{0.545} & 0.502 \ci{0.486}{0.519} \\
\textsc{laya-ml} & 0.635 \ci{0.620}{0.652} & 0.604 \ci{0.588}{0.621} & 0.644 \ci{0.628}{0.660} & 0.631 \ci{0.615}{0.647} & 0.673 \ci{0.657}{0.688} & 0.521 \ci{0.505}{0.538} & 0.477 \ci{0.459}{0.493} \\
\textsc{von} & 0.852 \ci{0.840}{0.864} & 0.852 \ci{0.840}{0.864} & -- & 0.852 \ci{0.840}{0.864} & 0.852 \ci{0.840}{0.864} & 0.500 \ci{0.483}{0.517} & 0.500 \ci{0.483}{0.517} \\
Qwen2.5-1.5B & 0.500 \ci{0.479}{0.521} & 0.500 \ci{0.479}{0.521} & -- & 0.513 \ci{0.492}{0.534} & 0.511 \ci{0.490}{0.532} & 0.500 \ci{0.479}{0.521} & 0.508 \ci{0.487}{0.529} \\
Qwen2.5-7B & 0.789 \ci{0.770}{0.806} & 0.915 \ci{0.902}{0.927} & -- & 0.909 \ci{0.897}{0.921} & 0.835 \ci{0.820}{0.851} & 0.500 \ci{0.479}{0.522} & 0.516 \ci{0.495}{0.539} \\
Llama-3.1-8B & 0.638 \ci{0.618}{0.659} & 0.753 \ci{0.734}{0.772} & -- & 0.514 \ci{0.493}{0.536} & 0.500 \ci{0.478}{0.521} & 0.500 \ci{0.479}{0.522} & 0.503 \ci{0.482}{0.525} \\
Mistral-7B & 0.723 \ci{0.705}{0.743} & 0.858 \ci{0.842}{0.872} & -- & 0.877 \ci{0.862}{0.890} & 0.867 \ci{0.852}{0.881} & 0.510 \ci{0.490}{0.531} & 0.492 \ci{0.471}{0.514} \\
\bottomrule
\end{tabular}
}
\caption{PolicyBench-XF accuracy by label condition. The decision rule appears only in the option
definition. Chance is 0.500. \textbf{label only} and \textbf{neither} are controls establishing
that the definition is the sole information source.}
\label{tab:policy}
\end{table}

\begin{table}[t]\centering\small
\noindent\resizebox{\linewidth}{!}{\begin{tabular}{lrrr}
\toprule
model & arbitrary $-$ aligned & misleading $-$ aligned & arbitrary exch. $-$ arbitrary \\
\midrule
\textsc{laya-td} & $+0.1511$ \ci{+0.1377}{+0.1646} & $+0.0942$ \ci{+0.0754}{+0.1129} & $+0.0017$ \ci{-0.0014}{+0.0051} \\
\textsc{laya-en} & $+0.0776$ \ci{+0.0631}{+0.0923} & $-0.2618$ \ci{-0.2851}{-0.2386} & $-0.0063$ \ci{-0.0114}{-0.0014} \\
\textsc{laya-ml} & $-0.0042$ \ci{-0.0217}{+0.0131} & $+0.0376$ \ci{+0.0143}{+0.0611} & $-0.0003$ \ci{-0.0154}{+0.0149} \\
\textsc{von} & $+0.0000$ \ci{+0.0000}{+0.0000} & $+0.0000$ \ci{+0.0000}{+0.0000} & $+0.0000$ \ci{+0.0000}{+0.0000} \\
Qwen2.5-1.5B & $+0.0133$ \ci{+0.0076}{+0.0195} & $+0.0111$ \ci{-0.0305}{+0.0529} & $+0.0796$ \ci{+0.0667}{+0.0933} \\
Qwen2.5-7B & $+0.1203$ \ci{+0.1019}{+0.1395} & $+0.0464$ \ci{+0.0195}{+0.0733} & $+0.0305$ \ci{+0.0210}{+0.0405} \\
Llama-3.1-8B & $-0.1239$ \ci{-0.1390}{-0.1081} & $-0.1381$ \ci{-0.1781}{-0.1000} & $+0.3230$ \ci{+0.3014}{+0.3448} \\
Mistral-7B & $+0.1531$ \ci{+0.1343}{+0.1724} & $+0.1438$ \ci{+0.1224}{+0.1652} & $-0.0285$ \ci{-0.0390}{-0.0181} \\
\bottomrule
\end{tabular}
}
\caption{Paired contrasts on PolicyBench-XF, bootstrap over items.}
\label{tab:contrasts}
\end{table}

For \textsc{laya-td}, accuracy is \dneutral{} \ci{\dneutrallo}{\dneutralhi} higher when the
options are called \code{A} and \code{B} than when they are named after the decision
($n=\dneutraln{}$),
and labels that point to the wrong class still give higher accuracy than correct ones, by
\dmislead{} \ci{\dmisleadlo}{\dmisleadhi}. The control cells are within seven points of chance, so the suite measures definition following
rather than any property of the labels. The gain does not depend on which uninformative label is
used: unrelated words, digits and letters all recover most of the gap, at \polunrelated{},
\poldigits{} and \polletters{} against \polcanonical{} for the meaningful names, and
\textsc{von} returns the same accuracy under all of them. The state-blindness column of Table~\ref{tab:mitigations} shows that \textsc{laya-td} returns its
no-state decision on \sbaligned{} of items under aligned labels against \sbarbitrary{} under
arbitrary ones, so a
meaningful label displaces the state rather than supplementing it. This accounts for aligned
labels being worse than arbitrary ones. It does not by itself predict the ordering of the
misleading condition, which for this checkpoint is higher than aligned; the strings
\code{escalate} and \code{hold} are not neutral with respect to the policies, and which way they
push depends on the policy.

\paragraph{The direction of the effect differs between checkpoints.}
All three \textsc{laya} checkpoints depend on the labels, but they do not respond to misleading
labels in the same way. For \code{laya-typed-decisions}, misleading labels give higher accuracy
than aligned ones (\dmislead{} \ci{\dmisleadlo}{\dmisleadhi}). For \code{laya}, they reduce
accuracy from \layaenali{} to \layaenmis{} (\layaenmislead{}
\ci{\layaenmisleadlo}{\layaenmisleadhi}). We therefore report the contrast between arbitrary and
aligned labels as the primary quantity, because it measures whether the labels affect the decision
at all without depending on the direction in which a particular checkpoint responds.

\subsubsection{An artifact of the uncontrolled design}
\label{sec:confound}
Our first version of PolicyBench gave the false branch the catch-all definition ``none of the
above applies''. In that version, exchanging which letter carries the policy changes accuracy by a
clear margin (Table~\ref{tab:contrasts_plain}), which would ordinarily be read as a preference for
one option position over another, but that reading does not survive the control. When both options carry definitions
of comparable length and specificity, the same contrast is indistinguishable from zero for
\textsc{laya-td} and \textsc{laya-ml} (Table~\ref{tab:contrasts}). It remains small but non-zero
for the other systems, which we return to in \S\ref{sec:lm}. What the first version measured was a preference for whichever option
carried the more specific definition. The contrast between arbitrary and aligned labels survives the control for the two systems
that are well above chance on this suite, and is larger under it: \dneutral{} for
\textsc{laya-td} and \qseventbneutral{} for the 7B readout, against \plainneutraltd{} and
\plainneutralseven{} on the
uncontrolled suite. We therefore report the controlled suite throughout. We
describe the artifact because the uncontrolled design is the one a reader is likely to write
first.

\subsection{Language-model readouts show the same effect}
\label{sec:lm}

If the constrained decision head were responsible for this effect, an ordinary language model
scoring the same option strings would not be expected to show it. It shows it. On
PolicyBench-XF
(Table~\ref{tab:policy}) the Qwen2.5-7B label-log-probability readout gains
\qseventbneutral{} \ci{\qseventbneutrallo}{\qseventbneutralhi} from replacing meaningful labels
with \code{A} and \code{B} ($n=2100$; Table~\ref{tab:contrasts}), an effect of the same kind and
comparable size to the \dneutral{} \ci{\dneutrallo}{\dneutralhi} we measure for \textsc{laya-td}
($n=\dneutraln{}$). Its
label-only and neither cells sit at chance, so on this suite it too is reading the definitions
and being displaced by the labels.

The 1.5B row should not be read as evidence that scale removes the effect. Qwen2.5-1.5B is at chance in \emph{every} condition of PolicyBench-XF, including the
ones where the task is solvable, so its near-null contrast is a capacity floor rather than
robustness: a model that cannot do the task cannot display a label effect. The 7B readout is the
first language model in our set competent enough for the effect to be visible, and it is also the
one that shows it most strongly.

The result is not specific to one pretraining run. We repeat the measurement with two further
language-model families, Llama-3.1-8B and Mistral-7B, using the same readout and the same items.
Mistral-7B shows the effect in the same direction and at a comparable size,
\mistralneutral{} \ci{\mistralneutrallo}{\mistralneutralhi} from replacing meaningful labels with
\code{A} and \code{B}. On the classification tasks of \S\ref{sec:channels}, where all three
families are accurate under ordinary labels, all three lose a large amount of accuracy when a
label contradicts its definition: \confdropqwen{}, \confdropllama{} and \confdropmistral{} for
Qwen2.5-1.5B,
Llama-3.1-8B and Mistral-7B respectively, every interval excluding zero.

Llama-3.1-8B is the exception on PolicyBench, and the state-blindness statistic of
Equation~\eqref{eq:blind} identifies why. Under arbitrary labels it answers one class on 98.6\% of
items, and under contradicting labels its state-blindness is \llamablind{}: it returns the same
decision for every input, so there is no label effect to measure. This is the capacity floor
described above for Qwen2.5-1.5B rather than a model that resists the effect, and we read nothing
from its contrast on that suite. Its classification rows, where it is accurate, behave like the
others.

The readout is also sensitive to which letter carries which definition even when nothing in the
request is contradictory: exchanging the two letters changes Qwen2.5-7B accuracy by
\qsevenperm{} \ci{\qsevenpermlo}{\qsevenpermhi} (Table~\ref{tab:contrasts}), where the same
contrast is exactly zero for \textsc{von}. On the classification tasks of \S\ref{sec:channels} both families lose accuracy when a label and
its definition disagree, though not by the same amount: \textsc{laya-td} falls by
\confdroptdabs{} and the 1.5B readout by \confdropqwenabs{}. The 7B readout was not run on those tasks.

\paragraph{The two scoring readouts agree; the generative readout depends on the suite.}
Table~\ref{tab:readouts} and Figure~\ref{fig:readouts} in Appendix~\ref{app:tables} compare the
three readouts of \S\ref{sec:readouts} on one fixed backbone,
on both PolicyBench suites, and two of its results hold on both. Scoring the option strings by
length-normalized log probability and scoring them by a single constrained token give results that
agree within their intervals at every cell, so the particular scoring rule makes no difference.
This also rules out the length normalization of Equation~\eqref{eq:logprob} as a cause, because
the first-token readout applies no normalization and behaves the same way.

The comparison between scoring and generation does \emph{not} hold across suites, and we report
this because the uncontrolled result is the more striking one. On the uncontrolled suite the two
scoring readouts fall to chance under misleading labels (\sevenscoreMis{} at 7B) while the
generative readout reaches \sevengenMis{}. On the controlled suite the same comparison reverses:
the scoring readout reaches \xfscoreMis{} and the generative readout \xfgenMis{}. The collapse to
chance is therefore a property of the uncontrolled design, in which the false branch carries the
catch-all definition, and not a general property of scoring readouts. It is the same artifact
described in \S\ref{sec:confound}.

We therefore do not claim that the readout is a second condition for the effect. What the
controlled suite supports is narrower: the choice of scoring rule does not matter, and the label
effect itself is present in a language-model readout that has no typed decision head at all
(\xfscoreAli{} under aligned labels against \xfscoreMis{} under misleading ones, and
\qseventbneutral{} from arbitrary labels). That is what bears on the attribution in
\S\ref{sec:von} and \S\ref{sec:intervention}.

\subsection{A model that is unaffected, and why}
\label{sec:von}

Table~\ref{tab:invariance} shows \textsc{von} answering one routing request under five
labelings, with probabilities read back by which definition each label carries. The request is
written in the PolicyBench format rather than drawn from the generated set, so that all five
labelings can be displayed together.

\begin{table}[ht]\centering\small
\begin{tabular}{lrrr}
\toprule
rendering & $P(\text{false def.})$ & $P(\text{true def.})$ & confidence \\
\midrule
canonical, \code{hold}=F, \code{escalate}=T & 0.085900 & 0.914100 & 0.8280 \\
swapped, \code{escalate}=F, \code{hold}=T & 0.085900 & 0.914100 & 0.8280 \\
letters, \code{A}=F, \code{B}=T & 0.085900 & 0.914100 & 0.8280 \\
letters permuted, \code{B}=F, \code{A}=T & 0.085900 & 0.914100 & 0.8280 \\
unrelated, \code{alpha}=F, \code{beta}=T & 0.085900 & 0.914100 & 0.8280 \\
\midrule
\multicolumn{4}{l}{\small spread across renderings: $P(\text{true def.})$ = 0.0e+00} \\
\bottomrule
\end{tabular}

\caption{\textsc{von} answering one routing request under five labelings of the same question.
Probabilities are read back by which definition each label carries. The returned values are
identical to six decimal places in every row.}
\label{tab:invariance}
\end{table}

The spread across renderings is $0$ to machine precision. With both definitions replaced by the
same placeholder, \textsc{von} returns exactly $\{0.5, 0.5\}$. Across the seven policies its
accuracy under aligned, arbitrary and misleading labels is identical to four decimal places, and
its label-only and neither cells sit at exactly $0.5000$, so it takes nothing at all from the
labels.

\begin{figure}[t]\centering
\setlength{\tabcolsep}{4pt}\small
\resizebox{\linewidth}{!}{%
\begin{tabular}{@{}l l@{}}
\toprule
\multicolumn{2}{@{}l}{\textbf{Request}\quad options \code{hold} and \code{escalate}, with the
routing policy written in the definition of \code{escalate}} \\
\midrule
\textsc{laya} & \texttt{[state]\ [instructions]} \\
              & \colorbox{red!12}{\texttt{hold}}\texttt{: none of the above applies} \\
              & \colorbox{red!12}{\texttt{escalate}}\texttt{: the customer is on the Enterprise plan and asks for money back} \\
\addlinespace
\textsc{von}  & \texttt{[state]\ [instructions]} \\
              & \texttt{none of the above applies} \\
              & \texttt{the customer is on the Enterprise plan and asks for money back} \\
\bottomrule
\end{tabular}}
\caption{What reaches the encoder for the same request. \textsc{laya} prefixes each definition
with its option label (shaded), so the label competes with the definition the caller wrote.
\textsc{von} passes only the definitions and uses the label solely as a key for returning the
answer. \S\ref{sec:intervention} removes this prefix from \textsc{laya} and shows that the behavior
reported in \S\ref{sec:channels} and \S\ref{sec:policyresults} goes with it.}
\label{fig:mechanism}
\end{figure}

Figure~\ref{fig:mechanism} shows what reaches the encoder in each case, and the cause is one
line in each code base, shown together below. \textsc{laya} concatenates the
label in front of the definition, so both enter the encoder. \textsc{von} packs only the
definition; there the label is a dictionary key used to retrieve the definition and to name the
answer, and it reaches the model only as a fallback when no definition was supplied.

\noindent\begin{minipage}{\linewidth}
\begin{lstlisting}[language=Python]
# laya/common.py:120  -- the label is concatenated in front of the definition
return [str(k) if v is None or v == "" else "%s: %s" % (k, render_criterion(v))
        for k, v in crit.items()]

# von/backends/option_marker_backend.py:633  -- only the definition is packed
desc = q.criteria.get(opt)
descriptions.append(desc.strip() if desc else opt.strip())
packed_text = model.pack_sequence(state_text, q.instructions, descriptions)
\end{lstlisting}
\end{minipage}

\textsc{von} on its own would be weak evidence about the decision head, because a model that
never receives the label cannot display a label effect; it shows that such a head \emph{can} be
label-invariant, not that the head is irrelevant. The intervention in \S\ref{sec:intervention}
is what settles this: it removes the label from \textsc{laya}'s input while leaving its decision
head, weights and calibration untouched, and the effect disappears. The design also carries a cost, because \textsc{von} never receives
the label, it cannot use a label that is informative, and a caller who supplies short or missing
definitions receives a uniform distribution.

\subsection{Adding and removing the label}
\label{sec:intervention}
\begin{table}[ht]\centering\small
\noindent\resizebox{\linewidth}{!}{\begin{tabular}{llrrrr}
\toprule
suite & model & aligned & arbitrary & misleading & arbitrary $-$ aligned \\
\midrule
PolicyBench-XF & \\textsc{laya-td} & 0.742 \ci{0.728}{0.756} & 0.893 \ci{0.883}{0.903} & 0.836 \ci{0.824}{0.849} & $+0.1511$ \ci{+0.1377}{+0.1646} \\
 & \\textsc{laya-td}, label removed & 0.891 \ci{0.881}{0.902} & 0.891 \ci{0.881}{0.902} & 0.891 \ci{0.881}{0.902} & $+0.0000$ \ci{+0.0000}{+0.0000} \\
 & \\textsc{laya-en} & 0.791 \ci{0.778}{0.804} & 0.869 \ci{0.857}{0.880} & 0.530 \ci{0.513}{0.546} & $+0.0776$ \ci{+0.0631}{+0.0923} \\
 & \\textsc{laya-en}, label removed & 0.853 \ci{0.841}{0.864} & 0.853 \ci{0.841}{0.864} & 0.853 \ci{0.841}{0.864} & $+0.0000$ \ci{+0.0000}{+0.0000} \\
 & \\textsc{laya-ml} & 0.635 \ci{0.620}{0.652} & 0.631 \ci{0.615}{0.647} & 0.673 \ci{0.657}{0.688} & $-0.0042$ \ci{-0.0217}{+0.0131} \\
 & \\textsc{laya-ml}, label removed & 0.626 \ci{0.610}{0.642} & 0.626 \ci{0.610}{0.642} & 0.626 \ci{0.610}{0.642} & $+0.0000$ \ci{+0.0000}{+0.0000} \\
 & \\textsc{von} & 0.852 \ci{0.840}{0.864} & 0.852 \ci{0.840}{0.864} & 0.852 \ci{0.840}{0.864} & $+0.0000$ \ci{+0.0000}{+0.0000} \\
 & \\textsc{von}, label added & 0.875 \ci{0.864}{0.886} & 0.908 \ci{0.899}{0.918} & 0.834 \ci{0.822}{0.847} & $+0.0333$ \ci{+0.0246}{+0.0423} \\
\addlinespace
classification & \\textsc{laya-td} & 0.849 \ci{0.825}{0.873} & 0.797 \ci{0.770}{0.824} & 0.136 \ci{0.113}{0.158} & $-0.0516$ \ci{-0.0720}{-0.0324} \\
 & \\textsc{laya-td}, label removed & 0.807 \ci{0.780}{0.832} & 0.807 \ci{0.780}{0.832} & 0.807 \ci{0.780}{0.832} & $+0.0000$ \ci{+0.0000}{+0.0000} \\
 & \\textsc{laya-en} & 0.867 \ci{0.844}{0.890} & 0.812 \ci{0.784}{0.838} & 0.124 \ci{0.102}{0.146} & $-0.0551$ \ci{-0.0768}{-0.0336} \\
 & \\textsc{laya-en}, label removed & 0.831 \ci{0.806}{0.855} & 0.831 \ci{0.806}{0.855} & 0.831 \ci{0.806}{0.855} & $+0.0000$ \ci{+0.0000}{+0.0000} \\
 & \\textsc{laya-ml} & 0.792 \ci{0.765}{0.820} & 0.753 \ci{0.724}{0.782} & 0.218 \ci{0.190}{0.247} & $-0.0399$ \ci{-0.0612}{-0.0204} \\
 & \\textsc{laya-ml}, label removed & 0.750 \ci{0.720}{0.779} & 0.750 \ci{0.720}{0.779} & 0.750 \ci{0.720}{0.779} & $+0.0000$ \ci{+0.0000}{+0.0000} \\
 & \\textsc{von} & 0.851 \ci{0.827}{0.875} & 0.851 \ci{0.827}{0.875} & 0.851 \ci{0.827}{0.875} & $+0.0000$ \ci{+0.0000}{+0.0000} \\
 & \\textsc{von}, label added & 0.857 \ci{0.833}{0.880} & 0.851 \ci{0.826}{0.875} & 0.228 \ci{0.200}{0.256} & $-0.0058$ \ci{-0.0180}{+0.0072} \\
\addlinespace
\bottomrule
\end{tabular}
}
\caption{The rendering changed in both directions, on both suites. For each \textsc{laya}
checkpoint the patched row renders an option as its definition alone instead of
\code{"\{label\}: \{definition\}"}. For \textsc{von} the patched row does the reverse and prepends
the label. Weights, decision head and calibration are unchanged within each pair, and the items
are the same.}
\label{tab:intervention}
\end{table}

Comparing \textsc{laya} with \textsc{von} compares two systems that differ in their weights, their
training data and their rendering at once. To isolate the rendering, we patched it in both
directions. In the three \textsc{laya} checkpoints we changed the expression quoted above so that
an option is rendered as its definition alone. In \textsc{von} we did the reverse, prepending each
option's label to its definition, which is what \textsc{laya} does. Each patch is a change to one
expression; nothing else about either system was touched.

Table~\ref{tab:intervention} gives the result, and removing the label makes all three \textsc{laya}
checkpoints exactly invariant: aligned, arbitrary and misleading labels return the same accuracy
on both suites, and the contrast between arbitrary and aligned labels is $+0.0000$ with a
bootstrap interval of $[+0.0000, +0.0000]$, because the three requests have become the same input.
The collapse under conflicting labels goes with it. On the classification tasks
\code{laya-td} falls to \chanswapnames{} when a label contradicts its definition, and the patched
model stays at \nolabcls{}.

Adding the label to \textsc{von} produces the effect. Its contrast on PolicyBench moves from
exactly zero to \vonlabpold{} \ci{\vonlabpoldlo}{\vonlabpoldhi}, and on the classification tasks it
acquires the same collapse as the \textsc{laya} checkpoints, falling from \vonchan{} under aligned
labels to \vonlabclsmis{} when a label contradicts its definition. A model that is invariant to
option labels by construction becomes label-sensitive when the label is written into its input,
with no change to its weights.

The two directions together identify the rendering as the cause. The effect is removed from three
checkpoints of one system by deleting the label, and created in a second, independently trained
system by inserting it.

Removing the label is not free, and it moves accuracy in opposite directions on the two suites.
On PolicyBench, where the labels carry no information about the policy, removing them raises
\code{laya-td} from \polcanonical{} to \nolabpol{}. On the classification tasks, where the class
names are informative, it lowers accuracy from \chancanonical{} to \nolabcls{}. This is the trade
that separates \textsc{von} from \textsc{laya} in \S\ref{sec:channels}, reproduced inside a single
system.

\subsection{A two-call test}
\label{sec:diagnostic}
The measurements above reduce to a test that a caller can run in two calls. Ask the same question
twice, changing only the option labels and leaving the definitions unchanged, then compare the two
returned distributions. If they are identical, the model is in the position of \textsc{von}: the
definitions are its entire input and need to be written so that they are sufficient on their own.
If they differ, the model is in the position of \textsc{laya}, and the mitigations of
\S\ref{sec:mitigations} apply. The test requires no gold labels, no held-out data and no access to
the weights, so it can be run against a hosted model.

\subsection{Negated questions}
\label{sec:polarity}
\begin{figure}[t]\centering
\includegraphics[width=0.92\linewidth]{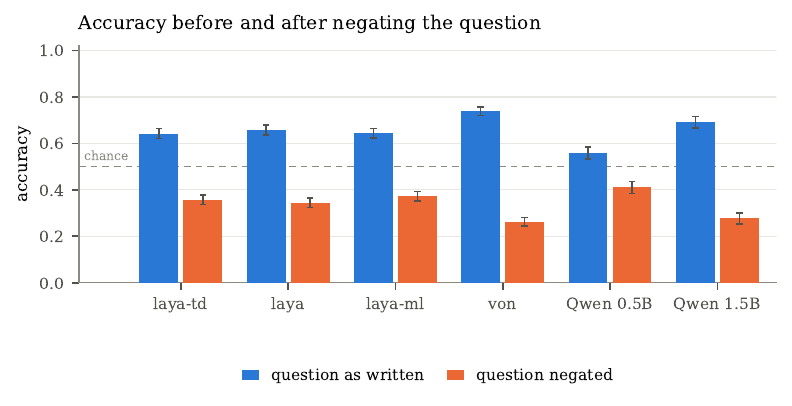}
\caption{Accuracy on yes/no questions before and after negating the question, pooled over seven
tasks. The returned probability is re-oriented for the negated form, so a model that reads the
negation would score the same in both conditions. Every model instead falls from above chance to
below it by about the same margin, which is what happens when the returned probability does not
change and only the correct answer has been inverted. Exact values and flip rates are in
Table~\ref{tab:polarity} of Appendix~\ref{app:tables}.}
\label{fig:polarity}
\end{figure}

Prefixing a yes/no question with ``Is it false that'' changes the decision on \polflip{}
\ci{\polfliplo}{\polfliphi} of items for \textsc{laya-td}. The flip rate lies between \polflipml{} and \polflip{} across the three
\textsc{laya} checkpoints. The returned probability is
essentially unchanged by the negation. Accuracy
moves from \polcan{} to \polneg{}, and $\polcan{} + \polneg{} \approx 1$, which is what results
when the returned probability is unchanged and only the gold label has been inverted. An explicit
instruction to attend to the polarity of the question leaves this untouched (\polwarn{}).

Negated phrasing is common in ordinary requests. A developer who writes a check as ``is it false that the request is authorized'' rather than ``is
the request unauthorized'' receives an inverted answer, reported with normal confidence.

\paragraph{Invariance to labels does not imply that the question is read.}
\textsc{von} is the most accurate of the four typed models on these tasks (\vonpolcan{}) and is
invariant to the option labels, but it changes its decision on \vonpolflip{}
\ci{\vonpolfliplo}{\vonpolfliphi} of negated items, which is within the range we measure for the
checkpoints that do use labels. The two failures are independent of each other, because the first depends on whether the
label is written into the model input, which the implementer controls; the second is unaffected by
that choice and is present in every typed model we tested. Adopting a label-invariant model
therefore removes one of these failures and leaves the other unchanged.

\paragraph{Comparison with the language-model baseline.}
The Qwen2.5-1.5B label-log-probability readout changes its decision on \lmpolflip{} of negated
items, against \polflipml{} to \polflip{} for the typed heads. It is also far from reading polarity
correctly, but it is affected less, and it is the kind of system that a typed decision model is
usually offered as a cheaper replacement for. Combined with \S\ref{sec:channels}, we do not find
the typed interface to be better than this baseline on either property we measured: the two are
comparable on label dependence and the typed models are worse on polarity. The reported case for typed decision models rests on latency and cost \citep{bench}. We did not
measure latency or cost, and nothing here bears on that case.

\subsection{Behavior of the confidence scores}
\label{sec:confidence}
\begin{table}[t]\centering\small
\noindent\resizebox{\linewidth}{!}{\begin{tabular}{llrrr}
\toprule
model & condition & accuracy & mean confidence & AUROC of confidence \\
\midrule
\textsc{laya-td} & aligned & 0.8487 & 0.4022 & 0.8123 \\
 & misleading & 0.1357 & 0.3397 & 0.2433 \\
\addlinespace
\textsc{laya-en} & aligned & 0.8667 & 0.6139 & 0.8236 \\
 & misleading & 0.1236 & 0.5852 & 0.1940 \\
\addlinespace
\textsc{laya-ml} & aligned & 0.7923 & 0.6830 & 0.7292 \\
 & misleading & 0.2185 & 0.5944 & 0.2871 \\
\addlinespace
\textsc{von} & aligned & 0.8511 & 0.5487 & 0.8055 \\
 & misleading & 0.8511 & 0.5487 & 0.8055 \\
\addlinespace
\bottomrule
\end{tabular}
}
\caption{Whether a model's own confidence separates its correct from its incorrect answers, by
condition, on fixed-label choice tasks. An AUROC below $0.5$ means the signal is actively
misleading about its own errors.}
\label{tab:confidence}
\end{table}

Under ordinary labels, the confidence score returned by \code{laya-typed-decisions} separates its
correct from its incorrect answers at AUROC \confauccan{}, so it carries usable information about
which decisions to trust. When a label and its definition disagree, and accuracy falls to
\confaccswap{}, the same score reaches AUROC \confaucswap{}. A value below $0.5$ means the
ranking is reversed, so the decisions the model reports as most confident are the ones more likely
to be wrong. \textsc{laya-en} behaves the same way (Table~\ref{tab:confidence}).

\textsc{von} provides the control, with an AUROC of \vonconfauccan{} under ordinary labels and
\vonconfaucswap{} under misleading ones, which are equal because its decisions in the two
conditions are identical. The confidence score degrades only in the conditions where the labels
change the decision, which indicates that the degradation follows the corrupted decision rather
than the manipulation itself.

This bears directly on the deployment pattern recommended throughout this literature, in which a
typed decision is accepted when confidence is high and escalated to a stronger model when it is
low \citep{judge}. Under a corrupted rendering that gate routes the wrong items: it forwards the
cases the model got right and accepts the ones it got wrong. A confidence threshold cannot detect
this failure, which is why we propose the rendering-disagreement check of \S\ref{sec:mitigations}
instead.

\section{Mitigations}
\label{sec:mitigations}
This section is written for a reader deciding what to do about the effect in their own system.
The order to work through is: run the test in \S\ref{sec:diagnostic} to find out whether the
labels are affecting your decisions at all; if they are, apply M1 where the task allows it; if the
labels have to keep their meaning, apply M2 and gate the remaining decisions with M4. Each
mitigation below is reported with what it was worth in our measurements rather than recommended on
principle.
\begin{table}[t]\centering\small
\noindent\resizebox{\linewidth}{!}{\begin{tabular}{llrrrr}
\toprule
model & condition & raw & prior-corrected & $\Delta$ & state-blind \\
\midrule
\textsc{laya-td} & aligned & 0.7420 & 0.7811 & $+0.0391$ & 0.6340 \\
 & arbitrary \texttt{A}/\texttt{B} & 0.8931 & 0.8509 & $-0.0423$ & 0.5240 \\
 & misleading & 0.8363 & 0.7877 & $-0.0486$ & 0.4546 \\
\addlinespace
\textsc{laya-en} & aligned & 0.7911 & 0.7163 & $-0.0749$ & 0.4900 \\
 & arbitrary \texttt{A}/\texttt{B} & 0.8689 & 0.8286 & $-0.0403$ & 0.5003 \\
 & misleading & 0.5297 & 0.5934 & $+0.0637$ & 0.9383 \\
\addlinespace
\textsc{von} & aligned & 0.8520 & 0.5000 & $-0.3520$ & 0.6029 \\
 & arbitrary \texttt{A}/\texttt{B} & 0.8520 & 0.5000 & $-0.3520$ & 0.6029 \\
 & misleading & 0.8520 & 0.5000 & $-0.3520$ & 0.6029 \\
\addlinespace
Qwen2.5-7B & aligned & 0.7886 & 0.8633 & $+0.0748$ & 0.7114 \\
 & arbitrary \texttt{A}/\texttt{B} & 0.9090 & 0.9167 & $+0.0076$ & 0.5157 \\
 & misleading & 0.8352 & 0.9071 & $+0.0719$ & 0.6581 \\
\addlinespace
\bottomrule
\end{tabular}
}
\caption{Null-state prior correction on PolicyBench-XF. \emph{state-blind} is the fraction of
decisions equal to the decision the same rendering makes with no state at all.}
\label{tab:mitigations}
\end{table}

\begin{table}[t]\centering\small
\noindent\resizebox{\linewidth}{!}{\begin{tabular}{llrrrrr}
\toprule
model & second rendering & flag rate & precision & recall & acc.\ unflagged & coverage \\
\midrule
\textsc{laya-td} & arbitrary \texttt{A}/\texttt{B} & 0.1900 & 0.8977 & 0.6611 & 0.8921 & 0.8100 \\
 & arbitrary \texttt{alpha}/\texttt{beta} & 0.1386 & 0.9031 & 0.4850 & 0.8458 & 0.8614 \\
\addlinespace
\textsc{laya-en} & arbitrary \texttt{A}/\texttt{B} & 0.1954 & 0.6988 & 0.6539 & 0.9102 & 0.8046 \\
 & arbitrary \texttt{alpha}/\texttt{beta} & 0.2214 & 0.6013 & 0.6375 & 0.9028 & 0.7786 \\
\addlinespace
\textsc{von} & arbitrary \texttt{A}/\texttt{B} & 0.0000 & -- & 0.0000 & 0.8520 & 1.0000 \\
 & arbitrary \texttt{alpha}/\texttt{beta} & 0.0000 & -- & 0.0000 & 0.8520 & 1.0000 \\
\addlinespace
Qwen2.5-7B & arbitrary \texttt{A}/\texttt{B} & 0.1995 & 0.8019 & 0.7568 & 0.9358 & 0.8005 \\
 & arbitrary \texttt{alpha}/\texttt{beta} & 0.1890 & 0.8338 & 0.7455 & 0.9336 & 0.8110 \\
\addlinespace
\bottomrule
\end{tabular}
}
\caption{Running the decision under a second rendering and flagging disagreement. The detector
needs two renderings that behave differently, a precondition the state-blindness statistic
checks.}
\label{tab:detector}
\end{table}

\paragraph{M1: arbitrary labels.} Name the options \code{A}, \code{B} and so on, and put the
whole of the rule in the definitions. This is worth \dneutral{} on PolicyBench-XF. On the
classification tasks it costs accuracy instead, because there the class name is itself a good
predictor of the class. The choice therefore depends on the task, and the test in
\S\ref{sec:diagnostic} is how to decide which case applies.

\paragraph{M2: null-state prior correction.} Send the same question once with an empty state,
which gives the distribution the model returns when it has no evidence, then divide it out of
every subsequent answer and renormalize:
\begin{equation}
\tilde{p}(o_i \mid s) \;\propto\; \frac{p(o_i \mid s)}{\,p(o_i \mid \varnothing)^{\alpha}},
\qquad \alpha = 1 .
\label{eq:prior}
\end{equation}
Equation~\eqref{eq:prior} is applied once per question and the empty-state call is cached, so the
cost is negligible. This is contextual calibration \citep{zhao2021} applied to a typed request.

Table~\ref{tab:mitigations} shows that the correction helps the models whose errors come from a
label-induced preference, and harms the model that does not have one. It is worth \mittdcan{} for
\textsc{laya-td} under aligned labels and \mitqseven{} for the Qwen2.5-7B readout, and it costs
accuracy once the labels are already arbitrary. For \textsc{von} it is actively damaging,
reducing accuracy to exactly $0.5000$ in all three conditions. The reason is specific to how the
correction is defined. The empty-state call removes the state but keeps the option definitions,
and the definitions are the whole of \textsc{von}'s input, so $p(o_i \mid \varnothing)$ already
contains the asymmetry that \textsc{von} uses to answer. Dividing it out removes the signal rather
than a bias. The correction should therefore be applied only to models that the test in
\S\ref{sec:diagnostic} places in the label-using regime.

\paragraph{M3: an instruction to ignore the labels.} Adding a sentence to the instructions
stating that the labels are arbitrary identifiers and that the definitions determine the answer is
worth \mthree{} \ci{\mthreelo}{\mthreehi} for \textsc{laya-td} on PolicyBench-XF, against
\dneutral{} for relabeling the options, so it is not a substitute for M1.

\paragraph{M4: two-rendering disagreement.} For \textsc{laya-td}, flagging the items on which the
aligned and arbitrary renderings disagree raises accuracy on the items that remain from
\detbase{} to \detacc{}, at \detcov{} coverage and \detprec{} precision, for one extra call and no
training. Where the task permits arbitrary labels, M1 is the better choice: it reaches
\polletters{} on the same model at full coverage. M4 is for the case where the labels must keep
their meaning.

The flag rate of this procedure is the same quantity as the test in \S\ref{sec:diagnostic}, and
the two are reported together in Table~\ref{tab:detector}. For \textsc{von} the flag rate is
$0.0000$, because its two renderings produce identical decisions. A flag rate of zero indicates
that the procedure is unnecessary for that model, not that the procedure has failed, so a caller
can decide what to do from this one number. The procedure is uninformative in one other case: if a
model cannot perform the task at all, both renderings produce near-constant output, they agree
with each other, and both are wrong. The neither control separates that case from the case of a
model that is reading the definitions.

\paragraph{Summary of the mitigations.} Where the task allows it, use arbitrary labels and state
the rule in the definitions; this is the largest effect we measured and it costs one edit to the
request. Where the labels have to carry meaning, because they are themselves good predictors of
the class, apply the cached null-state correction instead. Where individual decisions are worth
checking, run a second rendering and review the items on which the two disagree. Do not use a
confidence threshold on its own for this purpose: under a label that contradicts its definition
the confidence score ranks errors in the wrong direction (\S\ref{sec:confidence}), so the
threshold forwards the decisions the model got right and accepts the ones it got wrong.

\section{Limitations}
\label{sec:limitations}
We did not measure Jev, the hosted commercial model, so our claims cover open-weight
implementations of its interface and language-model readouts; for Jev itself we rely on
\citet{sun}. The mechanism
claim is scoped accordingly: we exhibit the responsible line in two open code bases and cannot
inspect a closed one.

PolicyBench is synthetic and templated, which buys exact ground truth, exact class balance and
immunity to contamination at the cost of naturalness; the natural-task suite covers the other side
of that trade.

The misleading condition presents a request whose label and definition disagree, which a reader
could call under-specified. The argument therefore rests on the arbitrary condition against the
arbitrary-exchanged condition, where nothing is contradictory, and on the contrast between
arbitrary and aligned labels, where no request contains a conflict at all.

The definitions are our own wording, although the control cells establish that they are legible rather than
decorative, but different phrasing would move the absolute numbers.

\code{laya-multilingual} is near chance throughout PolicyBench and we draw no conclusion from its
contrast.

\section{Conclusion}
A typed decision model's \code{criteria} field is a contract: the caller writes what each option
means, and expects the returned probability to be a posterior over those meanings. Across
open-weight models we find that one of the four implementations satisfies this and three do not.
Architecture, training objective and parameter count do not separate the two groups. What
separates them is whether an option's label is written into the model's input alongside its
definition. We show this directly rather than by comparison: removing the label from the three
systems that have the effect eliminates it, and adding it to the system that does not have it
creates it, with no change to any weight. The change costs nothing in latency, and practitioners
can determine which regime their own model is in with two calls.

The two evaluation gaps identified by \citet{audit} are related to each other. Without an
invariance test over option names, the effect reported here is not visible at all, and without a
language-model baseline it cannot be separated from the architecture that the models introduce.
Reporting both is what allowed us to locate the behavior in the prompt rendering rather than in
the decision head.

\section*{Ethics statement}
This work evaluates publicly released model weights on public datasets and on synthetic data we
generate. It involves no human subjects, no personal data and no new data collection. PolicyBench
tickets are generated from templates and describe fictional customers. The failure modes we report
are already present in deployed systems and the paper gives detection and mitigation procedures
for them, so we judge the risk of publication to be low relative to the benefit of disclosure.
We evaluate third-party models under their licenses (Apache-2.0 for the typed decision models) and
name specific checkpoints only to make the results reproducible, not to rank vendors.

\section*{Reproducibility statement}
All models are open-weight and all datasets are public, so no vendor API key is needed to
reproduce any number in this paper. Every reported measurement is generated from cached per-item predictions by
\code{src/make\_tables.py} and inserted through a macro, so a rerun of the grid updates the
manuscript. The only literals in the prose are values that are exactly zero or exactly one half,
quoted as such. Section~\ref{sec:method} specifies the surface variants, the state ablations and the
bootstrap procedure; Section~\ref{sec:policybench} specifies the PolicyBench generator, whose
ground truth is computed from the generating attributes. Code, the PolicyBench generator, and the
raw per-item model outputs for every grid cell are released.

\bibliographystyle{iclr2027_conference}
\bibliography{refs}

\appendix
\section{Supplementary tables}
\label{app:tables}

\begin{figure}[htb]\centering
\includegraphics[width=\linewidth]{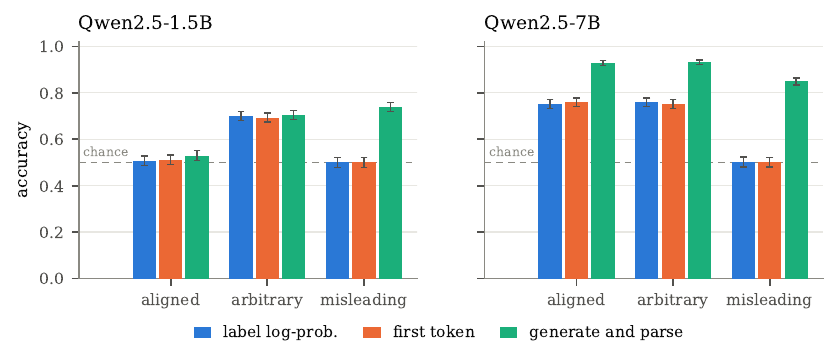}
\caption{Accuracy on PolicyBench for three readouts over one fixed backbone, by label condition.
Shown on the uncontrolled suite, where the gap between the scoring and generative readouts is
largest; \S\ref{sec:lm} reports what happens under the control. Error bars are bootstrap 95\%
intervals over items.}
\label{fig:readouts}
\end{figure}

\begin{table}[ht]\centering\small
\noindent\resizebox{\linewidth}{!}{\begin{tabular}{lrrrrr}
\toprule
model & aligned & label only & arbitrary \texttt{A}/\texttt{B} & neither & misleading \\
\midrule
\textsc{laya-td} & 0.849 \ci{0.825}{0.873} & 0.856 \ci{0.832}{0.879} & 0.797 \ci{0.770}{0.824} & 0.569 \ci{0.537}{0.604} & 0.136 \ci{0.113}{0.158} \\
\textsc{laya-en} & 0.867 \ci{0.844}{0.890} & 0.862 \ci{0.839}{0.885} & 0.812 \ci{0.784}{0.838} & 0.552 \ci{0.519}{0.587} & 0.124 \ci{0.102}{0.146} \\
\textsc{laya-ml} & 0.792 \ci{0.765}{0.820} & 0.759 \ci{0.729}{0.788} & 0.753 \ci{0.724}{0.782} & 0.559 \ci{0.526}{0.594} & 0.218 \ci{0.190}{0.247} \\
\textsc{von} & 0.851 \ci{0.827}{0.875} & 0.510 \ci{0.477}{0.544} & 0.851 \ci{0.827}{0.875} & 0.510 \ci{0.477}{0.544} & 0.851 \ci{0.827}{0.875} \\
Qwen2.5-0.5B & 0.701 \ci{0.664}{0.738} & 0.664 \ci{0.624}{0.701} & 0.679 \ci{0.643}{0.714} & 0.482 \ci{0.441}{0.521} & 0.476 \ci{0.434}{0.513} \\
Qwen2.5-1.5B & 0.755 \ci{0.722}{0.787} & 0.768 \ci{0.735}{0.798} & 0.776 \ci{0.742}{0.807} & 0.547 \ci{0.506}{0.585} & 0.207 \ci{0.175}{0.240} \\
Llama-3.1-8B & 0.776 \ci{0.742}{0.807} & 0.736 \ci{0.701}{0.769} & 0.662 \ci{0.624}{0.698} & 0.550 \ci{0.512}{0.589} & 0.452 \ci{0.412}{0.491} \\
Mistral-7B & 0.833 \ci{0.803}{0.861} & 0.785 \ci{0.752}{0.817} & 0.791 \ci{0.758}{0.823} & 0.485 \ci{0.447}{0.524} & 0.529 \ci{0.490}{0.567} \\
\bottomrule
\end{tabular}
}
\caption{Exact values for Figure~\ref{fig:channels}: what each channel contributes, on the
fixed-label choice tasks (SST-2, RTE, CB; $n=\chann{}$). The ordinal task is excluded because a
\code{score} question has no label channel separate from its definition.}
\label{tab:channels}
\end{table}

\begin{table}[ht]\centering\small
\noindent\resizebox{\linewidth}{!}{\begin{tabular}{lrrrr}
\toprule
model & as written & negated & negated + warning & decision flip rate \\
\midrule
\textsc{laya-td} & 0.642 \ci{0.621}{0.663} & 0.358 \ci{0.337}{0.378} & 0.355 \ci{0.334}{0.376} & 0.983 \ci{0.977}{0.988} \\
\textsc{laya-en} & 0.658 \ci{0.637}{0.679} & 0.345 \ci{0.324}{0.366} & 0.342 \ci{0.321}{0.362} & 0.982 \ci{0.975}{0.987} \\
\textsc{laya-ml} & 0.644 \ci{0.623}{0.665} & 0.372 \ci{0.351}{0.394} & 0.369 \ci{0.348}{0.390} & 0.966 \ci{0.958}{0.974} \\
\textsc{von} & 0.738 \ci{0.719}{0.756} & 0.263 \ci{0.244}{0.281} & 0.269 \ci{0.250}{0.287} & 0.971 \ci{0.964}{0.979} \\
Qwen2.5-0.5B & 0.559 \ci{0.534}{0.585} & 0.411 \ci{0.385}{0.436} & 0.425 \ci{0.400}{0.451} & 0.933 \ci{0.920}{0.946} \\
Qwen2.5-1.5B & 0.692 \ci{0.668}{0.716} & 0.278 \ci{0.253}{0.302} & 0.354 \ci{0.330}{0.380} & 0.797 \ci{0.777}{0.818} \\
\bottomrule
\end{tabular}
}
\caption{Exact values for Figure~\ref{fig:polarity}, with the decision flip rate of
Equation~\eqref{eq:flip} in the final column.}
\label{tab:polarity}
\end{table}

\begin{table}[ht]\centering\small
\noindent\resizebox{\linewidth}{!}{\begin{tabular}{llrrrrrr}
\toprule
& & \multicolumn{3}{c}{uncontrolled suite} & \multicolumn{3}{c}{controlled suite (XF)} \\
\cmidrule(lr){3-5}\cmidrule(lr){6-8}
backbone & readout & aligned & arbitrary \texttt{A}/\texttt{B} & misleading & aligned & arbitrary \texttt{A}/\texttt{B} & misleading \\
\midrule
Qwen2.5-1.5B & label log-prob. & 0.507 \ci{0.486}{0.529} & 0.700 \ci{0.680}{0.720} & 0.500 \ci{0.479}{0.521} & 0.500 \ci{0.479}{0.521} & 0.513 \ci{0.492}{0.534} & 0.511 \ci{0.490}{0.532} \\
 & first token & 0.511 \ci{0.490}{0.533} & 0.693 \ci{0.673}{0.713} & 0.500 \ci{0.478}{0.521} & 0.500 \ci{0.479}{0.520} & 0.514 \ci{0.493}{0.535} & 0.511 \ci{0.490}{0.533} \\
 & generate and parse & 0.530 \ci{0.509}{0.552} & 0.704 \ci{0.685}{0.724} & 0.740 \ci{0.720}{0.759} & 0.591 \ci{0.570}{0.612} & 0.536 \ci{0.514}{0.557} & 0.523 \ci{0.501}{0.544} \\
\addlinespace
Qwen2.5-7B & label log-prob. & 0.751 \ci{0.733}{0.770} & 0.760 \ci{0.742}{0.778} & 0.501 \ci{0.480}{0.522} & 0.789 \ci{0.770}{0.806} & 0.909 \ci{0.897}{0.921} & 0.835 \ci{0.820}{0.851} \\
 & first token & 0.760 \ci{0.742}{0.779} & 0.752 \ci{0.734}{0.771} & 0.501 \ci{0.480}{0.522} & 0.800 \ci{0.783}{0.817} & 0.907 \ci{0.894}{0.919} & 0.832 \ci{0.816}{0.848} \\
 & generate and parse & 0.928 \ci{0.917}{0.939} & 0.932 \ci{0.921}{0.943} & 0.849 \ci{0.833}{0.864} & 0.930 \ci{0.920}{0.941} & 0.951 \ci{0.942}{0.960} & 0.793 \ci{0.776}{0.811} \\
\addlinespace
\bottomrule
\end{tabular}
}
\caption{One backbone, three readouts, on both PolicyBench suites. The two scoring readouts agree
throughout; the comparison with the generative readout does not hold across suites
(\S\ref{sec:lm}).}
\label{tab:readouts}
\end{table}

\begin{table}[ht]\centering\small
\noindent\resizebox{\linewidth}{!}{\begin{tabular}{lrrr}
\toprule
model & arbitrary $-$ aligned & misleading $-$ aligned & arbitrary exch. $-$ arbitrary \\
\midrule
\textsc{laya-td} & $+0.1201$ \ci{+0.1080}{+0.1317} & $+0.0912$ \ci{+0.0706}{+0.1120} & $+0.0660$ \ci{+0.0577}{+0.0749} \\
\textsc{laya-en} & $+0.2852$ \ci{+0.2669}{+0.3037} & $-0.0305$ \ci{-0.0603}{-0.0009} & $+0.0072$ \ci{-0.0017}{+0.0163} \\
\textsc{laya-ml} & $+0.0353$ \ci{+0.0171}{+0.0534} & $+0.0839$ \ci{+0.0580}{+0.1100} & $+0.0359$ \ci{+0.0169}{+0.0540} \\
\textsc{von} & $+0.0000$ \ci{+0.0000}{+0.0000} & $+0.0000$ \ci{+0.0000}{+0.0000} & $+0.0000$ \ci{+0.0000}{+0.0000} \\
Qwen2.5-1.5B & $+0.1928$ \ci{+0.1681}{+0.2171} & $-0.0073$ \ci{-0.0500}{+0.0352} & $-0.1766$ \ci{-0.2005}{-0.1529} \\
Qwen2.5-7B & $+0.0080$ \ci{-0.0090}{+0.0252} & $-0.2511$ \ci{-0.2867}{-0.2152} & $+0.1526$ \ci{+0.1314}{+0.1733} \\
\bottomrule
\end{tabular}
}
\caption{Contrasts on the uncontrolled PolicyBench suite, in which the false branch carries the
catch-all definition ``none of the above applies''. The third column is the artifact discussed in
\S\ref{sec:confound}; compare it with the same column of Table~\ref{tab:contrasts}.}
\label{tab:contrasts_plain}
\end{table}

\section{Tasks}
\label{app:tasks}
Table~\ref{tab:tasks} lists every task, its question type, the number of items used and the class
prevalence after any rebalancing. UNFAIR-ToS clause types are class-balanced before use because
their raw splits are about 97\% negative, where a model that always answers ``no'' reaches 0.97
accuracy while carrying no information.

\begin{table}[ht]\centering\small
\noindent\resizebox{\linewidth}{!}{\begin{tabular}{llrrrl}
\toprule
group & task & type & items & classes & class prevalence \\
\midrule
fixed-label choice & \code{sst2} & choice & 500 & 2 & 0.51, 0.49 \\
 & \code{rte} & choice & 277 & 2 & 0.53, 0.47 \\
 & \code{cb} & choice & 56 & 3 & 0.41, 0.50, 0.09 \\
\addlinespace
ordinal & \code{sst5} & score & 500 & 5 & 0.14, 0.31, 0.16, 0.23, 0.17 \\
\addlinespace
yes/no & \code{boolq} & boolean & 500 & 2 & 0.40, 0.60 \\
 & \code{sst2\_noul} & boolean & 500 & 2 & 0.51, 0.49 \\
 & \code{rte\_noul} & boolean & 277 & 2 & 0.47, 0.53 \\
 & \code{wic} & boolean & 500 & 2 & 0.50, 0.50 \\
 & \code{unfair\_tos\_0} & boolean & 76 & 2 & 0.50, 0.50 \\
 & \code{unfair\_tos\_1} & boolean & 76 & 2 & 0.50, 0.50 \\
 & \code{unfair\_tos\_2} & boolean & 76 & 2 & 0.50, 0.50 \\
\addlinespace
\bottomrule
\end{tabular}
}
\caption{Tasks used in the paper, with the number of items evaluated and the class prevalence
after any rebalancing.}
\label{tab:tasks}
\end{table}

\end{document}